\documentclass[letterpaper]{article} 
\usepackage[preprint]{aaai2027}  
\usepackage[hyphens]{url}  
\usepackage{graphicx} 
\usepackage{natbib}  
\usepackage{caption} 
\usepackage{amsmath}
\usepackage{amssymb}
\usepackage{booktabs}
\usepackage{multirow}

\newcommand{\evaf}{\textsc{EVAF}}
\newcommand{\rag}{\textsc{RAG}}
\newcommand{\lora}{\textsc{LoRA}}

\title{Memory Depth, Not Memory Access: Selective Parametric Consolidation for Long-Running Language Agents}

\author{
    Haoliang Han
}
\affiliations{
    Institute of Biomedical Strategy, China Pharmaceutical University\\
    \texttt{haolianghan1992@cpu.edu.cn}
}

\begin{document}

\maketitle

\begin{abstract}
Long-running language agents need more than memory access. Retrieval systems can
fetch past facts at query time, but they do not decide which experiences should
continue to shape behavior after the working context is unloaded. We study this
separate problem as \emph{memory depth}: durable goal-conditioned tendencies
written into a small parametric store. We introduce the loop-drift protocol, a
controlled stress test in which the retrieval index remains intact while working
context is unloaded and goal-conditioned behavior must persist under long-loop
interference. We evaluate \evaf, a surprise- and valence-gated \lora{}
consolidation mechanism. Across GPT-2 and TinyLlama, retrieval is strongest on
shallow factual recall (short-fact accuracy $0.956$--$0.973$), while \evaf{} is
strongest on goal persistence and post-unload recovery ($0.812$--$0.904$) with
only $2$--$3$ parametric writes per 200 events. Mechanism controls show that
selective consolidation factorizes into two controllable dimensions: selection
and actuation. Matched random gates isolate selection beyond sparse writing;
fixed-inner controls across GPT-2, TinyLlama, and Mistral-7B show that
inner-loop write strength is model-dependent; and a Mistral-7B matched-gate
inversion reveals asymmetric selection--actuation coupling under miscalibrated
actuation. Public Memora event streams serve as an external diagnostic,
exposing stale-memory invalidation as an unresolved boundary. Within this
probe, selective parametric consolidation supplies memory depth distinct from
and complementary to retrieval access.
\end{abstract}

\section{Introduction}

Long-horizon language agents accumulate histories that quickly exceed their
working context. The dominant engineering answer is retrieval: store past events
outside the model, retrieve a relevant subset, and condition generation on the
retrieved text \citep{lewis2020retrieval,packer2023memgpt,gutierrez2024hipporag}.
Retrieval is indispensable, but it answers a particular question: \emph{what can
be fetched?} It does not answer a complementary question: \emph{what should keep
shaping the agent's behavior even when no relevant text is in context?}

We call this second property \emph{memory depth}. A shallow memory is available
when a system retrieves or attends to it. A deep memory changes future behavior:
it persists through interference, survives context unload, and affects choices
without being reinserted as text. This distinction mirrors the motivation behind
complementary learning systems, where fast episodic stores and slow consolidating
stores serve different roles \citep{mcclelland1995complementary,kumaran2016learning}.
For language agents, the distinction is easy to blur. If a benchmark asks for an
old fact, retrieval should win. But a long-running assistant also needs durable
goals, preferences, and constraints that are not merely fetched facts.

We therefore study selective parametric consolidation under a controlled
loop-drift protocol. Each synthetic user stream contains stable goal events,
off-topic distractors, transient opposite requests, conflicts, sibling-user
contamination, and explicit factual notes. We probe multiple memory layers:
recent facts, old noisy facts, goal persistence, and post-unload goal recovery.
The key design is that retrieval memory is durable. Context unload clears the
working context, not the retrieval index. Thus an \evaf{} advantage on the goal
layer is not a trivial ``RAG forgot'' artifact.

Our mechanism, \evaf{}, uses a surprise-times-valence gate to admit only
behavior-relevant events into a small buffer. When
the buffer fills, a low-rank adapter is updated with replay and an L2 anchor.
This is not meant to replace retrieval: it targets the slower question of which
events should leave a persistent behavioral imprint.

The paper makes five contributions:

\begin{enumerate}
\item We formulate the memory-access versus memory-depth split for long-running
language agents and instantiate it in a controlled loop-drift protocol.
\item We show a depth flip across GPT-2 and TinyLlama: \rag{} wins shallow
facts, while \evaf{} wins goal persistence and post-unload recovery with far
fewer writes than naive continual \lora{}.
\item We show that selection is not only sparsity: a matched random gate with
the same write count loses to \evaf{} on GPT-2, while Mistral-7B exposes how
miscalibrated actuation can invert the same comparison.
\item We separate selection from actuation. Fixed-inner controls across GPT-2,
TinyLlama, and Mistral-7B show that write strength is a distinct,
model-dependent factor, and that over-actuation can degrade both persistence
and selectivity.
\item We use public Memora event streams as an external diagnostic. The result
is deliberately modest: \evaf{} is directionally positive but not significant
on stale-memory rejection, exposing deletion/update validity as future work.
\end{enumerate}

\section{Memory Access vs. Memory Depth}

We define a long-running memory stream as a sequence of events
$x_1,\ldots,x_T$. At evaluation time, a method may have access to an external
store, a parametric adapter, or both. We distinguish four probe layers.

\textbf{Shallow episodic access} asks for a recent explicit fact. This is the
natural strength of retrieval.
\textbf{Noisy episodic access} asks for an older fact after same-key
interference. Retrieval is still expected to be competitive.
\textbf{Parametric tendency} asks whether a stable goal continues to shape
behavior after long interference.
\textbf{Post-unload recovery} repeats the goal probe immediately after a context
unload, with retrieval memory still intact but working context cleared.

The target is not universal memory accuracy. The target is a more specific
tradeoff:
\[
  \begin{aligned}
  &\max\;\mathrm{GoalPersist}+\mathrm{PostUnload}\\
  &\text{s.t. low writes and bounded adapter drift.}
  \end{aligned}
\]
Short factual access is expected to be owned by retrieval. The contribution of
parametric consolidation is the goal-conditioned layer that remains active under
unload.

\section{Method}

\subsection{EVAF Selection Gate}

\evaf{} maintains a small write buffer. For each event $x_t$, the model computes
a surprise score $s_t$ from token negative log-likelihood and a valence score
$v_t$ from embedding similarity to the user's durable goal and preferences. The
write admission score is
\[
  g_t = \sigma(k_s(s_t-\tau_s))\cdot\sigma(k_v(v_t-\tau_v)).
\]
If $g_t>\tau_w$, the event enters the buffer. When the buffer reaches a fixed
size, the adapter is updated on the buffer plus replay from previous
consolidated events. The adapter is a \lora{} module \citep{hu2021lora}; replay
and an L2 anchor act as drift guards, following the broad continual-learning
intuition behind replay and elastic constraints
\citep{rolnick2019experience,kirkpatrick2017overcoming}.
We use a per-user median warmup for the surprise threshold, fixed gate slopes
($k_s=1$, $k_v=10$), $\tau_v=0.5$, and $\tau_w=0.5$ across seeds; full constants
are listed in the supplement.

\subsection{Actuation Controllers}

The original \evaf{} implementation coupled selection with a fixed inner-loop
write strength. Our later controls show that this is not enough: the same
selected buffer can help or hurt depending on how strongly it is written. We
therefore evaluate actuation separately from event selection.

\textbf{Fixed-inner controllers} are the primary diagnostic: they use the same
\evaf{} selection gate but vary only the number of inner \lora{} steps. We
report fixed-1, fixed-2, and fixed-3.

\textbf{Behavior-margin control} is included as an engineering proof of concept
rather than as a first-class mechanism claim. It mechanically extracts held-in
calibration pairs from the training event itself, never from evaluation stems,
and writes until the event-derived goal/preference margin moves by a target
amount, subject to KL and L2 caps. We report it because it offers a
behavior-level actuation controller that complements the fixed-inner diagnostic,
but its results are not load-bearing: the Mistral cross-seed variance is much
larger than for fixed-inner controls, and we do not use it for any 7B repair
claim.

\textbf{Routed EVAF+RAG} routes factual probes to retrieval and goal probes to
\evaf. It tests complementarity: retrieval for access, parametric consolidation
for depth.

\subsection{Mechanistic Intuition}

For a stream loss $\ell_t(\theta)$, a naive continual adapter applies updates on
nearly every event,
\[
  \Delta\theta_{\mathrm{naive}} \propto -\sum_{t=1}^{T}\nabla_\theta
  \ell_t(\theta).
\]
\evaf{} instead admits a sparse subset $A=\{t:g_t>\tau_w\}$ and updates
\[
  \Delta\theta_{\mathrm{evaf}} \propto -\sum_{t\in A}\nabla_\theta
  \ell_t(\theta) - \lambda(\theta-\theta_0).
\]
This should reduce drift when most stream events are distractors, but sparsity
alone is not the mechanism: if the admitted subset is random, the update can
still point toward transient, sibling, or off-topic gradients. The gate is useful
only if $g_t$ is correlated with behavior-relevant events. The matched-gate
ablation tests exactly this condition.

\subsection{Separable Factors, Coupled Dynamics}

Selection and actuation are separable control factors, but they are not
independent online dynamics. Selection drives actuation directly: the gate
decides whether a buffer is written. Actuation feeds back into future selection
indirectly: after the adapter changes, later surprise scores are computed under
the new model state.

This asymmetric feedback is visible in the Mistral-7B fixed-inner audit
(Table~\ref{tab:coupling}). Smaller inner steps trigger more future writes on
Mistral, consistent with weaker actuation leaving later similar events
surprising enough to pass the gate. The direction is monotonic across all four
Mistral seeds. On GPT-2 and TinyLlama, the same loop runs at much lower absolute
write counts (mean writes roughly 2--5), and the compressed range does not
produce a comparably clean monotonic signature; the coupling is most legible at
7B, where the gap between Fixed-1 and the default five-step controller spans
about six writes per user.

\begin{table}[t]
\centering
\small
\begin{tabular}{lcc}
\toprule
Method & Inner steps & Writes \\
\midrule
\evaf{} & 5 & 3.8 \\
Fixed-3 & 3 & 4.9 \\
Fixed-2 & 2 & 6.0 \\
Fixed-1 & 1 & 9.4 \\
\bottomrule
\end{tabular}
\caption{Asymmetric online feedback in the Mistral-7B four-seed fixed-inner
audit. Writes increase as inner-step strength decreases, indicating that
actuation changes future selection through model-state-dependent surprise.}
\label{tab:coupling}
\end{table}

\section{Loop-Drift Protocol}

Existing public long-memory benchmarks predominantly evaluate whether a system
can retrieve, update, or reason over stored information. They do not directly
isolate the post-unload setting we need here: the retrieval index remains
available, but the relevant text is absent from working context and the same
parametric write must continue to shape goal-conditioned behavior. Loop-drift is
therefore a controlled protocol rather than a leaderboard benchmark.

Each run contains 10 users and 200 events per user. Events are drawn from a
mixture of stable goal/preference reminders, distractors, transient opposite
requests, conflicts, sibling-user contamination, and scheduled factual notes.
We evaluate Frozen, Summary, \rag{}, Naive-LoRA, \evaf{}, and Routed
\evaf{}+\rag{}. GPT-2 and TinyLlama results are averaged over four seeds.
The \rag{} baseline stores all events in a durable embedding index and retrieves
the top three events by cosine similarity; context unload does not clear this
index.

The protocol is synthetic by design. The mechanism claim requires four
properties that are rarely controlled together in public memory benchmarks:
explicit interference, durable retrieval memory, an observable context unload,
and separate factual versus goal-conditioned probes. Synthetic control lets us
ask whether a method changed post-unload behavior rather than whether it found a
fact in an external store.

The protocol uses matched continuation scoring. Higher is better for
short-fact, long-fact, goal, and post-unload probes. Lower is better for
contamination and transient overwrite. Adapter cost is measured by write count
and \lora{} L2 drift. A write denotes one buffer-consolidation trigger, not one
gradient step; total inner updates equal writes times the controller's inner-loop
step count.

\section{Main Result: The Depth Flip}

\begin{figure*}[t]
\centering
\includegraphics[width=0.88\textwidth]{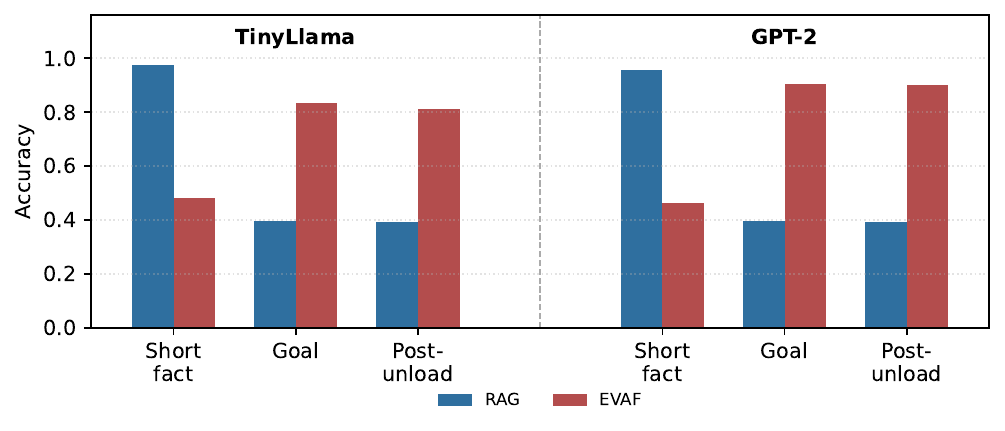}
\caption{The expected winner flips with memory depth. Retrieval is strongest on
shallow factual access, while \evaf{} is strongest on goal persistence and post-unload
recovery. Within each model, bars are grouped by probe depth; the dashed divider
separates TinyLlama from GPT-2. Values are four-seed means from the loop-drift
protocol.}
\label{fig:depth-split}
\end{figure*}

Table~\ref{tab:depth} and Figure~\ref{fig:depth-split} show the central result.
The expected winner changes with memory depth. \rag{} is strongest on recent explicit
facts, reaching $0.956$--$0.973$ short-fact accuracy. \evaf{} is near chance on
short facts, as expected: its gate is goal-conditioned and rejects off-topic
factual notes. But on the goal layer, \evaf{} is much stronger than \rag{}. On
TinyLlama, \evaf{} reaches $0.833$ goal persistence and $0.812$ post-unload
recovery, compared with \rag{} at $0.396$ and $0.394$. On GPT-2, \evaf{} reaches
$0.904$ and $0.900$, compared with \rag{} at $0.398$ and $0.394$.

\begin{table*}[t]
\centering
\small
\begin{tabular}{llcccccc}
\toprule
Model & Method & Short fact $\uparrow$ & Long fact $\uparrow$
& Goal $\uparrow$ & Post-unload $\uparrow$ & Writes $\downarrow$ & L2 $\downarrow$ \\
\midrule
\multirow{4}{*}{TinyLlama}
& \rag{} & \textbf{0.973} & \textbf{0.660} & 0.396 & 0.394 & 0.0 & 0.0 \\
& Naive-LoRA & 0.685 & 0.294 & 0.623 & 0.575 & 200.0 & 67.5 \\
& \evaf{} & 0.483 & 0.565 & \textbf{0.833} & \textbf{0.812} & 2.6 & 21.5 \\
& Routed & \textbf{0.975} & 0.634 & 0.773 & 0.750 & 2.5 & 21.2 \\
\midrule
\multirow{4}{*}{GPT-2}
& \rag{} & \textbf{0.956} & \textbf{0.733} & 0.398 & 0.394 & 0.0 & 0.0 \\
& Naive-LoRA & 0.733 & 0.330 & 0.642 & 0.650 & 200.0 & 119.3 \\
& \evaf{} & 0.463 & 0.558 & 0.904 & 0.900 & 2.4 & 29.0 \\
& Routed & 0.935 & 0.492 & \textbf{0.927} & \textbf{0.925} & 2.4 & 29.3 \\
\bottomrule
\end{tabular}
\caption{Depth split on the loop-drift protocol (four seeds, 10 users, 200
events). Retrieval owns shallow factual access; selective parametric
consolidation owns the context-unloaded goal layer. Naive-LoRA writes every
event and is costly without matching \evaf{} on the goal layer. Routed
\evaf{}+\rag{}
supports complementarity.}
\label{tab:depth}
\end{table*}

Routed \evaf{}+\rag{} is not the main mechanism, but it is an important sanity
check. On GPT-2, it recovers shallow factual access while slightly improving the
goal layer. On TinyLlama, it trades about six points of goal/post-unload
performance for near-perfect short-fact recall. This supports the division of
labor: retrieval and \evaf{} solve different memory problems, and routing can
combine them with model-dependent tradeoffs.

\section{Mechanism Controls}

The mechanism controls below are independent audit reruns, not extra rows from
the depth-split table. They use the same loop-drift generator and probe
families, but rerun the stream under controlled gate and actuation
counterfactuals; the margin controller additionally uses marker-style
training-only calibration pairs. Therefore, the actuation table should be read
within-run. Its ``Audit-5'' row is the original five-step \evaf{} controller in
that audit, not a second estimate of the Table~\ref{tab:depth} depth-split
number. The two runs use independent RNG streams and separate probe
instantiations; the actuation audit also includes the marker-style calibration
used by the Margin row. As a result, the same five-step controller can have
different absolute scores across tables (e.g., TinyLlama \evaf{} is $0.833$ in
Table~\ref{tab:depth}, while Audit-5 is $0.627$ in
Table~\ref{tab:actuation}). The within-table comparisons are the load-bearing
ones.

\subsection{Writing Everything Is Not Enough}

Naive-LoRA writes every event. It has 200 writes per stream and much higher L2
drift than \evaf{} (about $67$ on TinyLlama and $119$ on GPT-2 in the main
run), yet it does not cleanly solve the goal/post-unload layer. This shows that
the mechanism is not ``parametric memory'' alone. The question is which events
to admit and how strongly to write them. At 7B, indiscriminate writing becomes
actively harmful: in the Mistral matched-gate audit, Naive-LoRA collapses to
$0.333\pm0.047$ goal persistence, below the $0.500$ chance baseline, while
\evaf{} under the same uncorrected five-step actuation remains at
$0.362\pm0.092$. Writing without selection is therefore not merely inefficient;
at scale, it can move behavior in the wrong direction.

\subsection{Matched Gate: Selection Is Not Sparsity}

A natural objection is that \evaf{} works only because it writes sparsely. We
therefore compare against a random matched gate: the same write count and the
same online write dynamics, but random admitted events. This preserves the
coupled gate-actuation loop while isolating which events are selected.

On GPT-2, \evaf{} beats random matched gate on goal and post-unload in all four
seeds, with mean goal/post $0.790/0.763$ versus $0.590/0.619$. This establishes
that selection is not reducible to writing fewer events. TinyLlama is weak and
mixed ($0.625/0.581$ versus $0.596/0.569$), indicating that selection alone does
not determine the final behavior when actuation is poorly calibrated. Since
TinyLlama is larger than GPT-2 yet less clean in this matched-gate audit, the
selection signal is not a monotonic function of model scale under a fixed
five-step actuation rule.

Mistral-7B sharpens this point in the opposite direction. The fixed-inner audit
below shows that the original five-step actuation is miscalibrated at 7B; under
that wrong regime, the matched-gate comparison reverses, with random matched
writing ahead by $0.326$ goal and $0.369$ post-unload (\evaf{}
$0.362\pm0.092/0.312\pm0.092$; Random-Matched-Gate
$0.688\pm0.058/0.681\pm0.059$). This is not evidence that the \evaf{} gate
fails. It is a diagnostic signature of the asymmetric coupling in
Section~3.4: when actuation is miscalibrated, a behavior-targeted gate
concentrates the wrong update on goal-relevant events, while a random gate
spreads the same write strength across off-goal events. Notably, even in this
reversed regime, \evaf{} retains the lowest sibling contamination among
parametric methods ($0.787\pm0.041$ versus $0.825\pm0.046$ for random matched
gate and $0.870\pm0.046$ for Naive-LoRA; \rag{} is $0.995\pm0.004$). Selection
remains semantically active; what fails is its translation into goal behavior
under miscalibrated actuation. The fixed-inner audit below closes this bridge by
showing that the same gate recovers on Mistral when actuation is reduced.

\subsection{Actuation Controls}

Table~\ref{tab:actuation} reports the final fixed-inner audit. The original
five-step controller (Audit-5) over-actuates in these runs. Smaller fixed
controllers sharply reduce drift and improve goal/post-unload behavior, showing
that write strength is a separable control dimension rather than a constant
hyperparameter. There is no universal optimum: GPT-2 is strongest at Fixed-1 in
goal/post, TinyLlama reaches high goal under Fixed-1 and Margin but pays a
large contamination cost, and Mistral has stable recovery under Fixed-2 and
stronger but saturated behavior under Fixed-1. Behavior-margin control is kept
as a proof-of-concept actuation controller, but it is not the load-bearing
result. Contamination makes the tradeoff visible: Fixed-1 is best read as an
upper bound on goal movement, while Fixed-2 can offer a cleaner
selectivity--efficiency point, e.g., on GPT-2 it preserves high goal
($0.938$) with lower contamination ($0.203$) than Fixed-1.

\begin{table*}[t]
\centering
\small
\begin{tabular}{llccccc}
\toprule
Model & Method & Goal $\uparrow$ & Post $\uparrow$ & Contam. $\downarrow$
& Writes & L2 \\
\midrule
\multirow{5}{*}{TinyLlama}
& Audit-5 & 0.627 & 0.581 & \textbf{0.419} & 4.9 & 27.7 \\
& Fixed-1 & 0.929 & 0.925 & 0.868 & 4.8 & \textbf{13.3} \\
& Fixed-2 & 0.917 & 0.906 & 0.637 & 4.1 & 17.8 \\
& Fixed-3 & 0.702 & 0.663 & 0.547 & 4.4 & 22.4 \\
& Margin & \textbf{0.954} & \textbf{0.950} & 0.878 & 4.6 & 16.2 \\
\midrule
\multirow{5}{*}{GPT-2}
& Audit-5 & 0.885 & 0.881 & 0.337 & 2.4 & 29.2 \\
& Fixed-1 & \textbf{0.948} & \textbf{0.956} & 0.260 & 2.7 & \textbf{13.7} \\
& Fixed-2 & 0.938 & 0.925 & \textbf{0.203} & 2.2 & 18.6 \\
& Fixed-3 & 0.908 & 0.900 & 0.209 & 2.3 & 23.2 \\
& Margin & 0.910 & 0.906 & 0.247 & 2.2 & 16.6 \\
\bottomrule
\end{tabular}
\caption{Actuation audit. Selection and write strength are separable factors:
the same selection gate behaves differently under different inner-loop
strengths. Contamination is included because high goal/post performance can
come with selectivity cost. L2 should be compared within model, not across
model families. Audit-5 is the original five-step controller in this independent
audit rerun. Writes are consolidation triggers, not total gradient steps.}
\label{tab:actuation}
\end{table*}

Mistral-7B confirms the same factorization across four seeds. The original
five-step audit controller has goal/post
$0.354\pm0.075/0.306\pm0.080$. Fixed-2 recovers stable behavior
($0.796\pm0.035/0.775\pm0.041$), and Fixed-1 reaches
$0.919\pm0.081/0.938\pm0.095$. Fixed-1 contamination saturates at $1.000$ in
all four seeds, which indicates that this high-actuation regime crosses the
sibling probe's discriminative range rather than providing fine-grained
selectivity information. Margin is more variable on Mistral
($0.671\pm0.177/0.631\pm0.188$), so we do not use it as the 7B repair claim.
The 7B result is therefore not ``Fixed-1 solves scale,'' but a narrower
mechanistic conclusion: fixed five-step actuation is wrong, and inner-loop
strength is a separable, model-dependent control dimension.

\section{Boundary Diagnostic: Memora}

To probe the boundary of the mechanism claim, we test public Memora event
streams, which include memory mutation and stale evidence not present in the
loop-drift protocol. This is not an external validation win and not a SOTA
claim. It asks whether append-only selective consolidation also solves
stale-memory invalidation.

The full 10-persona diagnostic is weak but informative (Table~\ref{tab:memora}).
\evaf{} improves forgetting absence from $91/222$ to $95/222$, but the McNemar
test is not significant ($p=0.57$). A pre-registered high-mutation slice is
also non-significant. This means Memora should be read as a boundary: current
\evaf{} does not solve delete/update validity. Earlier negative-gradient
forgetting variants were unstable; validity gating or reconsolidation is the
more plausible future direction.

\begin{table}[t]
\centering
\small
\begin{tabular}{lccc}
\toprule
Setting & Frozen FA & \evaf{} FA & Test \\
\midrule
Full 10-persona & 91/222 & 95/222 & $p=0.57$ \\
High-mutation slice & 49/126 & 52/126 & $p=0.36$ \\
\bottomrule
\end{tabular}
\caption{Memora boundary diagnostic. Results are directionally positive but not
significant, so Memora is a public boundary result, not a victory benchmark.}
\label{tab:memora}
\end{table}

\section{Related Work}

\textbf{Retrieval and agent memory.}
Retrieval-augmented generation and agent memory systems fetch relevant past text
while keeping model weights fixed
\citep{lewis2020retrieval,packer2023memgpt,gutierrez2024hipporag}. Public
long-memory benchmarks such as LongMemEval and LoCoMo predominantly emphasize
conversational recall, temporal access, and knowledge updates
\citep{wu2024longmemeval,maharana2024evaluating}. Our point is not that
retrieval is weak. It is that these evaluations do not directly isolate the
post-unload setting in which retrieval remains available but behavior must
continue without reinserting the relevant text; the loop-drift protocol adds
that controlled probe.

\textbf{Neural memory and test-time fast weights.}
Recent architectures explicitly learn memory at test time. Titans introduces a
neural long-term memory module that complements attention
\citep{behrouz2025titans}, while In-Place TTT updates selected fast weights
during inference \citep{feng2026inplace}. Concurrent parametric-memory work
uses online \lora{} fast weights for self-evolving agents
\citep{ren2026scaling}, studies document \lora{} under KV-cache compression
\citep{zuo2026rethinking}, and quantifies exact \lora{} memory laws
\citep{xu2026lora}. Other recent agent-memory work studies retrieval-side
context validity \citep{yang2026ramem} or local Engram edits as an alternative
to per-user \lora{} contamination \citep{li2026engram}. These works are closest
in spirit to parametric memory. Our focus is different: we keep the base
architecture fixed and ask which agent-stream events should be selectively
consolidated into a small adapter so that behavior persists after context
unload.

\textbf{Continual learning and low-rank adaptation.}
Continual learning studies catastrophic forgetting, replay, and weight
regularization \citep{french1999catastrophic,kirkpatrick2017overcoming,rolnick2019experience}.
\lora{} provides a compact parametric store \citep{hu2021lora}. \evaf{} combines
these ingredients inside a selective event gate and evaluates whether the write
changes post-unload behavior, rather than reporting task accuracy alone.

\textbf{Complementary learning systems.}
The method is inspired by the broad CLS idea that fast episodic traces and slow
consolidating stores serve different functions
\citep{mcclelland1995complementary,kumaran2016learning,schapiro2017complementary}.
We use this as motivation only: \evaf{} is not a biological model. In particular,
stale-memory deletion remains unsolved and likely requires validity gating or
reconsolidation rather than anti-training.

\section{Scope and Limitations}

The paper establishes a narrow mechanism claim: memory depth can be probed by
post-unload goal-conditioned behavior, and selective parametric consolidation
factorizes into selection and actuation. It does not claim universal memory
accuracy, SOTA retrieval performance, robust paraphrase generalization, or
complete deletion/update validity.

First, the main protocol is synthetic. This is intentional: the mechanism claim
requires controlled interference, durable retrieval, and context unload. Memora
provides a public diagnostic, but it is not a significant win.

Second, heldout paraphrase is not robust. TinyLlama is weak and directional;
GPT-2 is not positive. We therefore claim durable held-in goal-conditioned
behavior, not broad semantic tendency generalization.

Third, the 7B result establishes the same actuation factorization, but not a
complete large-model memory system. Mistral still exhibits high contamination
under strong actuation, and Margin has high variance across seeds.

Fourth, high-actuation regimes can increase contamination. This is why we treat
selection and actuation as separable factors with online feedback, not as an
already-solved controller.

Finally, forgetting is not solved. Negative CE and anti-training are the wrong
object for stale-memory invalidation; future work should study validity-gated
reactivation and reconsolidation.

\section{Conclusion}

Memory access and memory depth are different problems. Retrieval is the right
tool for factual access. Selective parametric consolidation is a mechanism for
durable goal-conditioned behavior under interference and context unload. The
loop-drift protocol makes this split visible: \rag{} wins shallow facts, while
\evaf{} wins goal persistence and post-unload recovery with few writes and
bounded drift. The broader contribution is a mechanistic framing: consolidation
requires both semantic selection and calibrated actuation, and those factors are
separable controls with asymmetric online feedback. This gives future agent
memory work a more precise target than simply storing more text.

\bibliography{references}

\end{document}